\documentclass[sigconf]{acmart}

\AtBeginDocument{%
  }

\copyrightyear{2025}
\acmYear{2025}
\acmConference[]{arXiv}{preprint}{April 2025}
\makeatletter
\def\@conferenceinfo{}
\def\@acmConference{}
\makeatother

\renewcommand\footnotetextcopyrightpermission[1]{}
\usepackage{subfigure}
\usepackage{multirow} 
\usepackage{soul}

\newcommand{\zhiji}[1]{\textcolor{purple}{ Zhiji: #1}}

\begin{document}

\title[On the Effect of the Truncation Frequency in Multi-Objective Archiving]{When to Truncate the Archive?\\On the Effect of the Truncation Frequency in \\Multi-Objective Optimisation}



\author{Zhiji Cui}
\email{zxc990@student.bham.ac.uk}
\orcid{0009-0003-0639-8611}
\affiliation{%
  \institution{University of Birmingham}
  \city{Birmingham}
  \country{United Kingdom}
  \postcode{B15 2TT}
}

\author{Zimin Liang}
\orcid{0009-0005-6105-7777}
\email{zxl525@student.bham.ac.uk}
\affiliation{%
	\institution{University of Birmingham}
  \city{Birmingham} 
  \country{United Kingdom} 
  \postcode{B15 2TT}
}

\author{Lie Meng Pang}
\email{panglm@sustech.edu.cn}
\affiliation{%
  \institution{Southern University of Science and Technology}
  \city{Shenzhen}
  \country{China}
}

\author{Hisao Ishibuchi}
\email{hisao@sustech.edu.cn}
\affiliation{%
 \institution{Southern University of Science and Technology}
 \city{Shenzhen}
 \country{China}}

\author{Miqing Li}
\authornote{Corresponding author}
\orcid{0000-0002-8607-9607}
\email{m.li.8@bham.ac.uk}
\affiliation{%
  \institution{University of Birmingham}
  \streetaddress{Edgbaston}
  \city{Birmingham} 
  \country{United Kingdom} 
  \postcode{B15 2TT}
}


\begin{abstract}

Using an archive to store nondominated solutions found during the search of a multi-objective evolutionary algorithm (MOEA) is a useful practice. 
However, as nondominated solutions of a multi-objective optimisation problem can be enormous or infinitely many, it is desirable to provide the decision-maker with only a small, representative portion of all the nondominated solutions in the archive, thus entailing a truncation operation. 
Then, an important issue is when to truncate the archive. 
This can be done once a new solution generated, a batch of new solutions generated, or even using an unbounded archive to keep all nondominated solutions generated and truncate it later. 
Intuitively, the last approach may lead to a better result since we have all the information in hand before performing the truncation.
In this paper, we study this issue and investigate the effect of the timing of truncating the archive. 
We apply well-established truncation criteria that are commonly used in the population maintenance procedure of MOEAs (e.g., crowding distance, hypervolume indicator, and decomposition). 
We show that, interestingly, truncating the archive once a new solution generated tends to be the best, whereas considering an unbounded archive is often the worst. 
We analyse and discuss this phenomenon. 
Our results highlight the importance of developing effective subset selection techniques (rather than employing the population maintenance methods in MOEAs) when using a large archive.

\end{abstract}

\maketitle

\section{Introduction}

Multi-objective optimisation refers to an optimisation process that there are multiple objectives to be considered simultaneously. 
In multi-objective optimisation problems (MOPs), usually there does not exist one single optimal solution, but rather a set of trade-off solutions, called Pareto optimal solutions or a Pareto front in the objective space.
To tackle an MOP, optimisation algorithms, e.g., multi-objective evolutionary algorithms (MOEAs), are often designed to approximate the problem's Pareto front, such that the decision-maker can choose their preferred solution from the obtained approximation.

Recently, there has been an increasing interest in using a large (or even unbounded) archive to store the best solutions discovered throughout the search process of an MOEA. This can be attributed to 1) advancements in the computational and memory capabilities of modern computers, and 2) the inability of the final population of MOEAs to fully represent the best solutions discovered during the search process \cite{Laumanns2002,knowles_bounded_2004,Lopez2011,  li2019empirical}. This practice belongs to a broader topic, termed archiving~\cite{knowles_properties_2003, knowles_bounded_2004}, which studies the process of receiving new solutions from an optimiser, comparing them with the existing solutions in the archive, and determining which ones to keep or to discard.

In archiving, there are a variety of rules, criteria and strategies developed (see \cite{li2024multi,schutze2021archiving} for a survey). 
Some focus on making the archive non-deteriorate or converge over time, e.g., \cite{ hanne1999convergence, Laumanns2002b, knowles_properties_2003, schutze2007, Laumanns2011, Rudolph2016, zhang2025}. 
Some others directly consider well-established criteria used in the population update procedures of MOEAs for archiving \cite{knowles_bounded_2004,lopez2011sequential,li2019empirical,li2021our}, such as crowding distance \cite{Deb2002}, hypervolume indicator \cite{Beume2007} and decomposition criteria \cite{Zhang2007}, as population update can also be seen as an archiving process.
All of these methods involve an important issue -- when to maintain or truncate the archive.

Conventionally, the archive truncation can be performed either once a new solution arrives or a batch of new solutions arrive. The former corresponds to the case of updating the population once a new solution is generated (i.e., the $\mu+1$ evolution mode) in MOEAs such as SMS-EMOA \cite{Beume2007} and MOEA/D \cite{Zhang2007}, while the latter corresponds to the case of updating the population once a number of solutions are generated (i.e., the $\mu+\mu$ evolution mode) in MOEAs such as in NSGA-II \cite{Deb2002}, IBEA \cite{Zitzler2004} and NSGA-III \cite{Deb2014}. 
In addition, recently people tend to use an unbounded archive to store all nondominated solutions found during the search~\cite{Fieldsend2003, Ishibuchi2020,chen4902239updating,Rudolph2016,bezerra2019archiver,brockhoff2019benchmarking,li2024empirical,pang2020algorithm,bian2024archive}. 
Afterwards, the archive is truncated down to a small, but representative subset provided to the decision maker (as a large number of solutions can easily overwhelm them). 
Intuitively, one may think this approach is better since we have all the information in hand before performing the truncation. The $\mu+1$ approach that truncates the archive once every new solution arrives may be less promising due to the unknown about future input when the decision is made \cite{Lopez2011}.

In this paper, we study this issue to investigate the effect of the timing/frequency to perform the truncation to the archive. We consider the three approaches stated above, i.e., 1) truncating the archive once a new solution arrives (called \emph{immediate} truncation), 2) truncating the archive once a batch of new solutions arrive (called \emph{batch} truncation), and 3) truncating the archive after all solutions arrive (called \emph{unbounded} truncation). 
We apply well-established criteria that are commonly used in the population update
procedure of five MOEAs (NSGA-II, IBEA, SMS-EMOA, MOEA/D and NSGA-III), and would like to see how each one behaves under different truncation frequencies.
Interestingly, our experimental results show that in some MOEAs (e.g., SMS-EMOA and IBEA) the \emph{immediate} truncation tends to be the best,
whereas the \emph{unbounded} truncation is often the worst.
The reason for this occurrence is the use of the one-by-one solution removal strategy in population update of MOEAs.

The rest of the paper is organised as follows. Section 2 introduces preliminaries of multi-objective optimisation and multi-objective archiving. Section 3 is devoted to the experimental design, including the criteria used in the archive truncation and the sequences considered to feed the archives. Section 4 presents experimental results and discussions. 
Lastly, Section 5 concludes the paper.

\section{Preliminary} 

\subsection{Multi-objective optimisation problems}

Without loss of generality, let us consider a minimisation multi-objective optimisation problem (MOP) with $m$ objective functions $f(s) = (f_1(s),...,f_m(s))$ mapping a solution $s$ in a finite decision space $S$ to an objective space $Z$ ($Z \subseteq \mathbb{R}^m$).
For simplicity, we refer to an objective vector as a \emph{solution}.

Given two solutions $z, z'\in Z$, 
$z$ is said to \emph{(Pareto) dominate} $z'$ (denoted by $z \prec z'$) iff for all $i\in \{1,...,m\}$, $z_i \leq z'_i$ and $z \neq z'$.
Based on the dominance relation,
a solution $z\in Z$ is \emph{Pareto optimal} iff there does not exist another $z'\in Z$ such that $z'\prec z$. The set of Pareto optimal solutions is called the \emph{Pareto front}. Note that by convention the definition of Pareto optimality is based on solutions in the decision space, but here we extend it to the objective space as the archiving is typically related to the latter.


\subsection{Archiving}\label{sec:archiving}

The archiving process can be described as updating an archive $A$ by an \emph{input sequence}
$\mathcal{S}=\langle S^{(1)}$, $S^{(2)},\dots$, $S^{(T)}\rangle$,
which may be generated by a solution generator (e.g., an MOEA) iteratively \cite{li2024multi}.
At iteration $t$,
the generator may generate one or multiple solutions, i.e., $\forall t, |S^{(t)}| \geq 1$.
For the case that truncates the archive once a solution arrives, $|S^{(t)}|=1$ and $T=N$, where $N$ denotes the length of the sequence. For the case that truncates the archive after all solutions arrive, $|S^{(t)}|=N$ and $T=1$. For the case sitting between the two extremes (i.e., truncates the archive after a number of solutions arrive), $N>|S^{(t)}|>1$ and $N>T>1$ such that $|S^{(t)}| \times T = N$.

\section{Experimental Design}



As stated previously, we consider three approaches to truncate the archive. That is 
\begin{itemize}
    \item \emph{Immediate}: truncating the archive once a new solution arrives.
    \item \emph{Batch}: truncating the archive once $\mu$ solutions arrive.
    \item \emph{Unbounded}: truncating the archive after all solutions arrive.
\end{itemize}
Here, $\mu$ stands for the capacity of the archive, which resembles the ($\mu+\mu$) evolution mode in MOEAs, i.e., selecting $\mu$ solutions from $2\mu$ solutions. 

We consider well-known population truncation methods from five representative MOEAs. 
They are NSGA-II~\cite{Deb2002}, IBEA~\cite{Zitzler2004}, SMS-EMOA~\cite{Beume2007}, MOEA/D~\cite{Zhang2007} and NSGA-III~\cite{Deb2014}.
NSGA-II considers the Pareto dominance relation and crowdedness to perform the truncation. It selects solutions by using nondominated sorting to stratify solutions into layers and break ties within a layer by crowding distance.
IBEA and SMS-EMOA are indicator-based methods which use an indicator to measure the quality of the solutions in the archive. 
Specifically, IBEA uses a pairwise $\epsilon$-indicator~\cite{Laumanns2002} to assign fitness to each solution whereas SMS-EMOA uses the set-based hypervolume indicator~\cite{Zitzler1999} and compares the hypervolume contribution of each solution to the set.
MOEA/D and NSGA-III are decomposition-based methods which decompose the objective space into subspaces, each assigned with a directional reference point (also known as a reference vector).
Specifically, MOEA/D measures the solutions for each reference point by a scalarising function, and each directional vector only corresponds to one solution.
NSGA-III adopts the nondominated sorting mechanism of NSGA-II to select solutions while associating solutions to the nearest reference points, and prioritises the under-represented directional vectors (those with fewer associated solutions).
For simplicity, when referring to the truncation method of an MOEA we may use the MOEA itself instead.

\begin{figure}[tbp]
    \begin{center}
    \begin{tabular}{@{}c@{\hspace{0pt}}c@{\hspace{0pt}}c@{}}
        \includegraphics[scale=0.23]{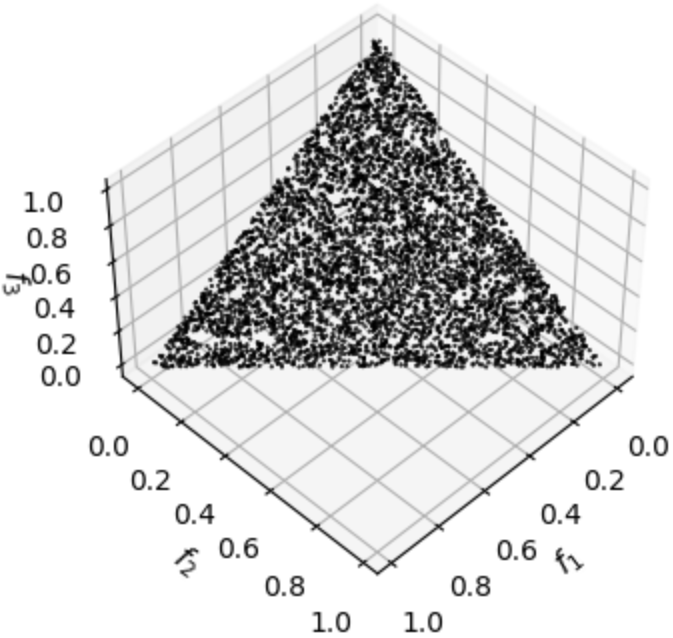} 
        \includegraphics[scale=0.23]{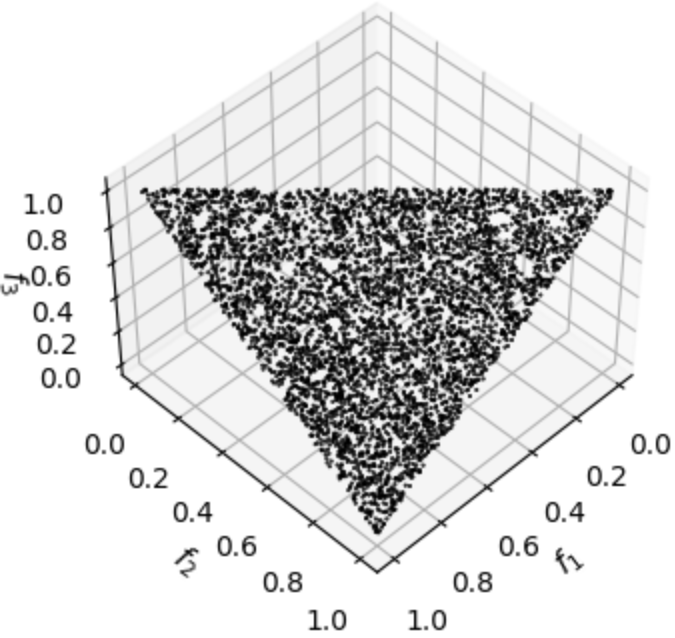}\\
       \makebox[4.2cm]{(a) \emph{simplex}} \makebox[4.2cm]{(c) \emph{inverted simplex}} \\
       
    \end{tabular}
    \end{center}
    \vspace{-10pt}
    \caption{Illustration of two types of test sequences generated from a simplex-shape Pareto front and an inverted simplex-shape Pareto front, respectively. Each type of sequence consists of 5,000 solutions randomly sampled from the (inverted) simplex.}
    \label{fig1}
\end{figure}

\begin{table*}[t]
\caption{IGD values (mean $\pm$ standard deviation) of the truncation methods from NSGA-II, SMS-EMOA, IBEA, MOEA/D and NSGA-III with the three truncation approaches. Statistical significance is measured by the Wilcoxon rank-sum test ($p < 0.05$), where the letter in the alphabetical order ($a$, $b$, $c$) indicates statistically significant superiority of one algorithm over another. That is, the approaches sharing the same letter have no statistical difference. For example, an algorithm labelled as ($a$) is statistically better than an algorithm labelled as ($b$), and neither of them is statistically different from an algorithm labelled as ($ab$).}
\footnotesize
\setlength{\tabcolsep}{5pt}
\label{tab:igd_results}
\begin{tabular}{@{}cllllll@{}}
\toprule
\textbf{\parbox[c]{1.5cm}{\centering Sequence}} & \multicolumn{1}{c}{\textbf{Truncation approach}} & \multicolumn{1}{c}{\textbf{NSGA-II}} & \multicolumn{1}{c}{\textbf{SMS-EMOA}} & \multicolumn{1}{c}{\textbf{IBEA}} & \multicolumn{1}{c}{\textbf{MOEA/D}} & \multicolumn{1}{c}{\textbf{NSGA-III}} \\
\midrule
\multirow{3}{*}{Simplex} 
& \multicolumn{1}{c}{\emph{immediate}} & 4.650e-02 ± 1.4e-03 $(a)$ & 3.798e-02 ± 1.4e-04 $(a)$ & 4.153e-02 ± 4.4e-04 $(a)$ & 3.718e-02 ± 0.0e+00 $(a)$ & 3.732e-02 ± 1.1e-05 $(ab)$ \\
& \multicolumn{1}{c}{\emph{batch}} & 4.835e-02 ± 1.9e-03 $(b)$ & 3.895e-02 ± 1.8e-04 $(b)$ & 4.150e-02 ± 5.3e-04 $(a)$ & 3.718e-02 ± 0.0e+00 $(a)$ & 3.732e-02 ± 7.6e-06 $(a)$\\
& \multicolumn{1}{c}{\emph{unbounded}} & 2.229e-01 ± 0.0e+00 $(c)$ & 3.911e-02 ± 9.7e-05 $(b)$ & 4.316e-02 ± 0.0e+00 $(b)$ & 3.718e-02 ± 0.0e+00 $(a)$ & 3.732e-02 ± 0.0e+00 $(b)$\\
\midrule
\multirow{3}{*}{\parbox[c]{1cm}{\centering Inverted\\simplex}}
& \multicolumn{1}{c}{\emph{immediate}}  & 4.638e-02 ± 1.1e-03 $(a)$ & 3.806e-02 ± 1.2e-04 $(a)$ & 4.673e-02 ± 5.4e-04 $(b)$ & 6.214e-02 ± 0.0e+00 $(a)$ & 4.856e-02 ± 8.3e-04 $(a)$\\
& \multicolumn{1}{c}{\emph{batch}} & 4.824e-02 ± 9.5e-04 $(b)$ & 3.904e-02 ± 1.9e-04 $(b)$ & 4.641e-02 ± 6.1e-04 $(ab)$ & 6.214e-02 ± 0.0e+00 $(a)$ & 4.938e-02 ± 9.8e-04 $(b)$\\
& \multicolumn{1}{c}{\emph{unbounded}} & 2.370e-01 ± 0.0e+00 $(c)$ & 3.959e-02 ± 6.0e-05 $(c)$ & 4.633e-02 ± 0.0e+00 $(a)$ & 6.214e-02 ± 0.0e+00 $(a)$ & 5.042e-02 ± 9.3e-04 $(c)$\\
\bottomrule
\end{tabular}
\vspace{1ex}

\end{table*}


We consider two types of sequences, each with 5,000 solutions, sampled randomly from a simplex Pareto front and an inverted simplex Pareto front respectively, shown in Figure~\ref{fig1}. For each type of sequences, we randomly shuffled 31 times, generating 31 different orders, as the order of the sequences may affect the archiving results \cite{li2021our}. We then apply the truncation methods to these sequences under the three truncation frequencies. We use the Wilcoxon rank-sum test to measure statistical soundness and consider the run with the median quality (evaluated by a quality indicator) for visualisation.  

We use the quality indicator IGD \cite{Coello2004a} to measure the quality of the truncation. The reference set used in IGD consists of 5,050 evenly distributed points on the (inverted) simplex, using the method in \cite{das1998normal}. Lastly, the archive capacity $\mu$ is set to 105 so that ideally a uniformly-distributed solution set in the simplex and inverted simplex plane can be obtained.

\section{Results}

In this section, we show the results of the truncation methods from the five MOEAs under the three truncation frequencies, through the IGD values and the plot of their solutions obtained. 
Table~\ref{tab:igd_results} gives IGD results (mean $\pm$ standard deviation). Here, the statistical significance is measured by the Wilcoxon rank-sum test ($p < 0.05$), where the letter in the alphabetical order ($a$, $b$, $c$) indicates statistically significant superiority of one algorithm over another. That is, the algorithms sharing the same letter have no statistical difference. For example, an algorithm labelled as ($a$) is statistically better than an algorithm labelled as ($b$), and neither of them is statistically different from an algorithm labelled as ($ab$).

\subsection{NSGA-II}

\begin{figure}[tbp]
    \begin{center}
    \begin{tabular}{@{}c@{\hspace{0pt}}c@{\hspace{0pt}}c@{}}
        \includegraphics[scale=0.155]{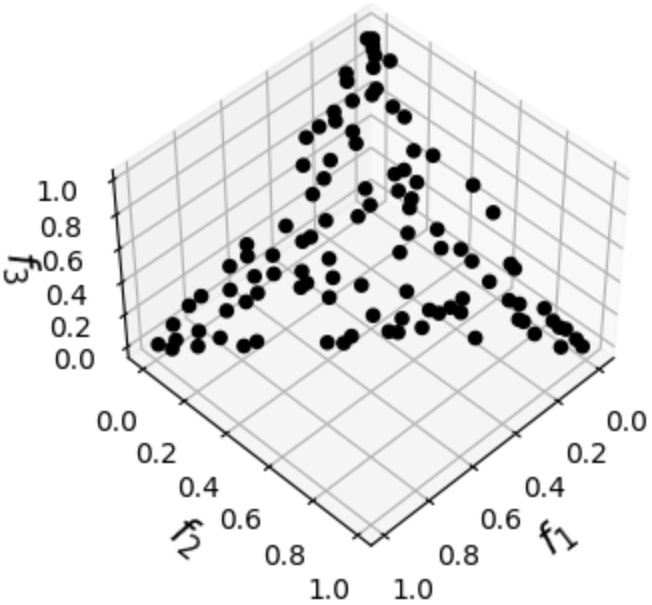} 
        \includegraphics[scale=0.155]{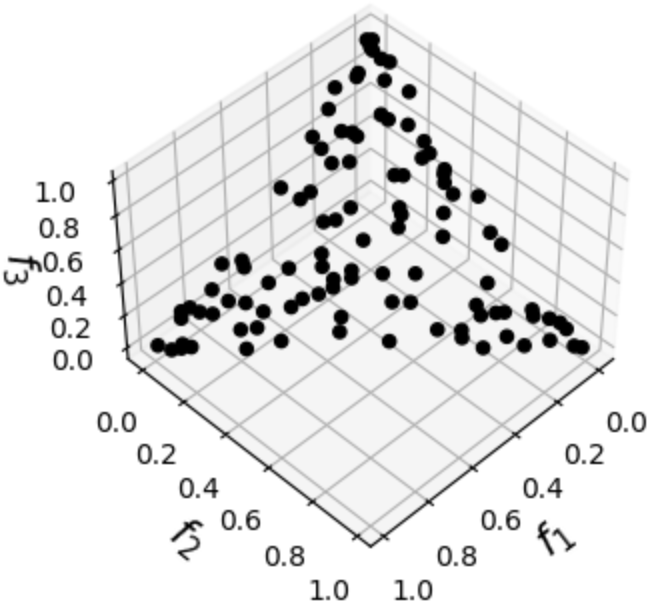} 
        \includegraphics[scale=0.155]{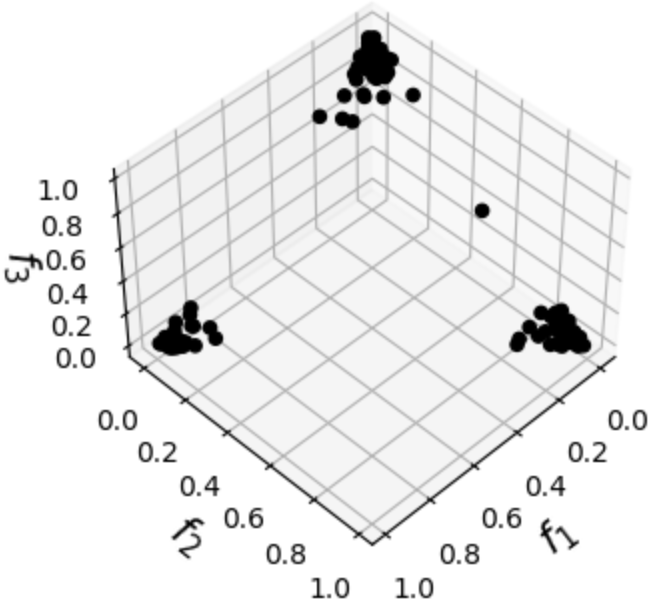} \\
       \makebox[2.765cm]{(a) \emph{immediate}} \makebox[2.765cm]{(b) \emph{batch}} \makebox[2.765cm]{(c) \emph{unbounded}} \\
       \makebox[2.765cm]{(IGD=0.0466)} \makebox[2.765cm]{(IGD=0.0480)} \makebox[2.765cm]{(IGD=0.2229)}
    \end{tabular}
    \end{center}
    \vspace{0pt}
    \begin{center}
    \begin{tabular}{@{}c@{\hspace{0pt}}c@{\hspace{0pt}}c@{}}
        \includegraphics[scale=0.155]{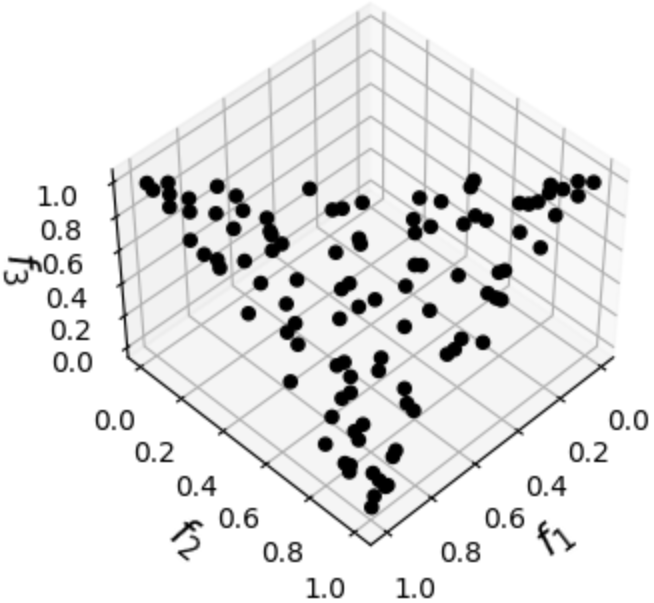} 
        \includegraphics[scale=0.155]{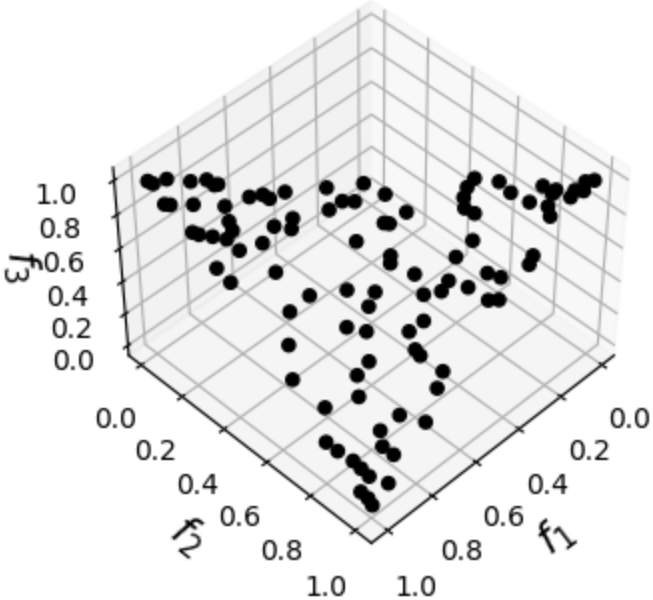} 
        \includegraphics[scale=0.155]{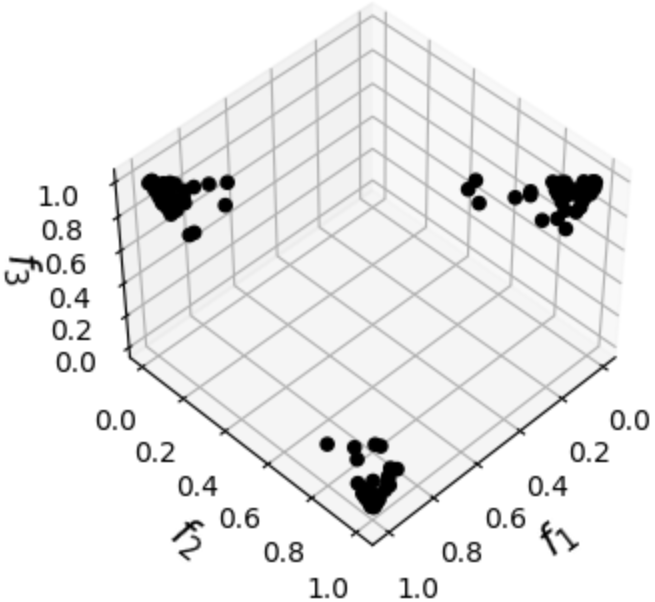} \\
       \makebox[2.765cm]{(a) \emph{immediate}} \makebox[2.765cm]{(b) \emph{batch}} \makebox[2.765cm]{(c) \emph{unbounded}} \\
       \makebox[2.765cm]{(IGD=0.0463)} \makebox[2.765cm]{(IGD=0.0482)} \makebox[2.765cm]{(IGD=0.2370)}
    \end{tabular}
    \end{center}
    \vspace{-10pt}
    \caption{Solutions (along with its IGD value) obtained by NSGA-II under the three truncation approaches on the simplex-shaped (top panel) and inverted simplex-shaped (bottom panel) sequences in the run with the median IGD value.}
    \label{fig2}
\end{figure}

NSGA-II employs the non-dominated sorting and crowding distance procedures in its population update. Note that here all solutions in the sequences are nondominated to each other, so the quality of solutions is only determined by crowding distance. As can be seen in Table~\ref{tab:igd_results}, the \emph{immediate} truncation approach obtains the best IGD value, followed by the \emph{batch} approach. The \emph{unbounded} approach has the worst result, falling far behind the other two (nearly an order of magnitude). This result can be confirmed in Figure~\ref{fig2}, where solutions under the three truncation frequencies in a run with the median IGD value are plotted. 

As we can see in the figure, the \emph{unbounded} truncation fails to cover the Pareto front, with their solutions clustering in the three corners. The reason for this occurrence is that NSGA-II uses a ``one-off'' way to perform the population update \cite{Deb2002}, i.e., it calculates the crowding distance of each solution at the beginning and then places $\mu$ solutions (here $\mu=105$) with the best crowding distance into the archive (i.e., effectively removing the (5000 - 105) worst solutions at one time). This practice, which does not update the crowding distance after a solution removed in the set, may lead to poor diversity (since the crowding distance is not accurate any more) \cite{Kukkonen2006a}. In the \emph{unbounded} truncation approach, this effect will become much more pronounced since there are $4,895$ solutions to be removed.

\begin{figure}[tbp]
    \begin{center}
    \begin{tabular}{@{}c@{\hspace{0pt}}c@{\hspace{0pt}}c@{}}
        \includegraphics[scale=0.155]{pictures/NSGA2-3.png} 
        \includegraphics[scale=0.155]{pictures/NSGA2-7.png} 
        \includegraphics[scale=0.155]{pictures/NSGA2-11.png} \\
       \makebox[2.765cm]{(a) \emph{immediate}} \makebox[2.765cm]{(b) \emph{batch}} \makebox[2.765cm]{(c) \emph{unbounded}} \\
       \makebox[2.765cm]{(IGD=0.0466)} \makebox[2.765cm]{(IGD=0.0480)} \makebox[2.765cm]{(IGD=0.2229)}
    \end{tabular}
    \end{center}
    \vspace{0pt}
    \begin{center}
    \begin{tabular}{@{}c@{\hspace{0pt}}c@{\hspace{0pt}}c@{}}
        \includegraphics[scale=0.155]{pictures/NSGA2-3.png} 
        \includegraphics[scale=0.155]{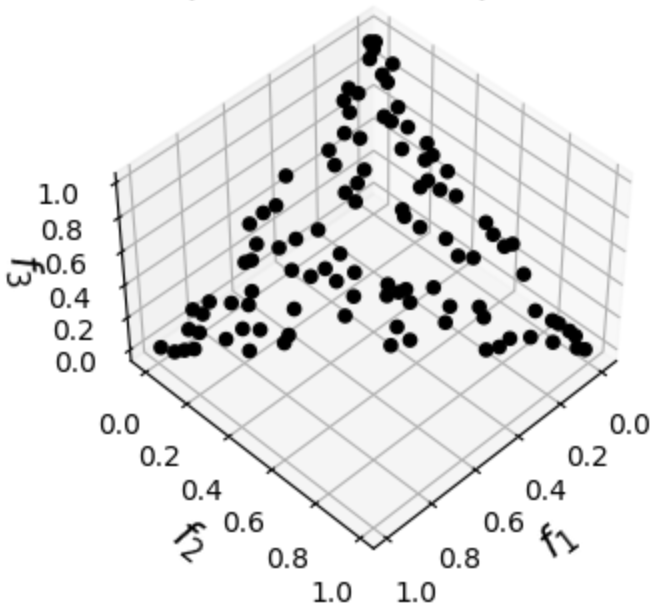} 
        \includegraphics[scale=0.155]{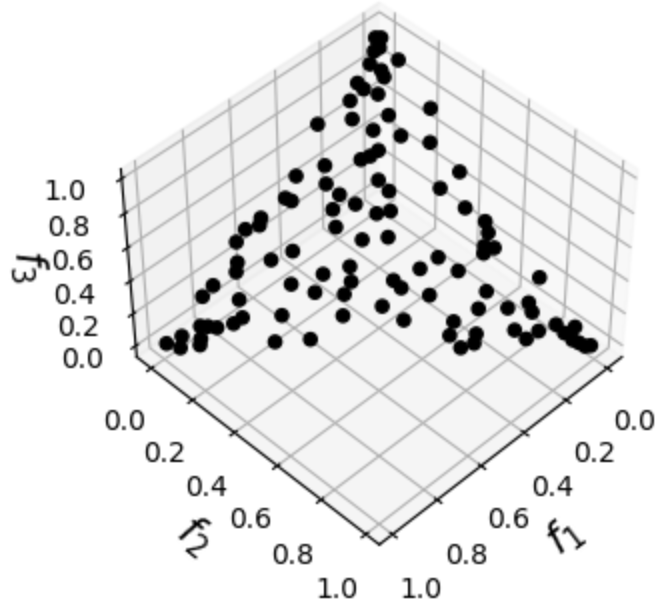} \\
       \makebox[2.765cm]{(a) \emph{immediate}} \makebox[2.765cm]{(b) \emph{batch}} \makebox[2.765cm]{(c) \emph{unbounded}} \\
       \makebox[2.765cm]{(IGD=0.0466)} \makebox[2.765cm]{(IGD=0.0466)} \makebox[2.765cm]{(IGD=0.0457)}
    \end{tabular}
    \end{center}
    \vspace{-10pt}
    \caption{The solutions obtained by the original NSGA-II (top panel) and the modified NSGA-II using the one-by-one truncation manner (bottom panel), along with their corresponding IGD values.}
    \label{fig3}
\end{figure}


 

To verify this, we modify the population update procedure of NSGA-II by performing the removal in the one-by-one manner (i.e., updating the crowding distance of the remaining solutions after one solution is removed). Figure~\ref{fig3} plots the solutions obtained by this modified method under the three truncation approaches (bottom panel). For comparison, the solutions obtained by the original NSGA-II are plotted as well (top panel). As can be seen, with this modification, the solutions of NSGA-II with the \emph{unbounded} approach can cover the whole Pareto front. There is also a small improvement of the IGD value for the \emph{batch} approach (0.0466 versus 0.0480). For the \emph{immediate} approach, the results are the same since it performs the truncation (i.e., updating the crowding distance) as long as a new solution arrives.  

Lastly, it is worth mentioning that regardless of the truncation approaches, uniformity of the solutions obtained by NSGA-II is poor. This is because the crowding distance fails to measure the crowdedness level of solutions in the space with three or more objectives -- an issue which has been frequently reported and studied in the area, theoretically \cite{zheng_runtime_2024} and empirically \cite{Kukkonen2006,Li2014a}.

\subsection{SMS-EMOA}

\begin{figure}[tbp]
    \begin{center}
    \begin{tabular}{@{}c@{\hspace{0pt}}c@{\hspace{0pt}}c@{}}
        \includegraphics[scale=0.155]{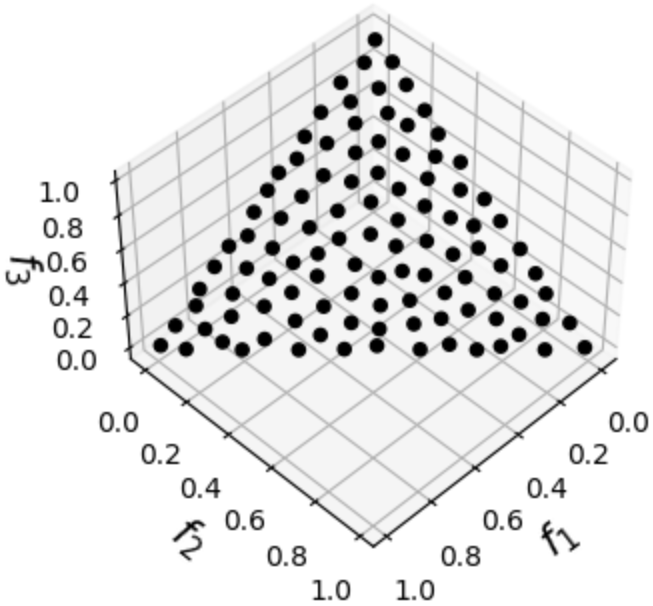} 
        \includegraphics[scale=0.155]{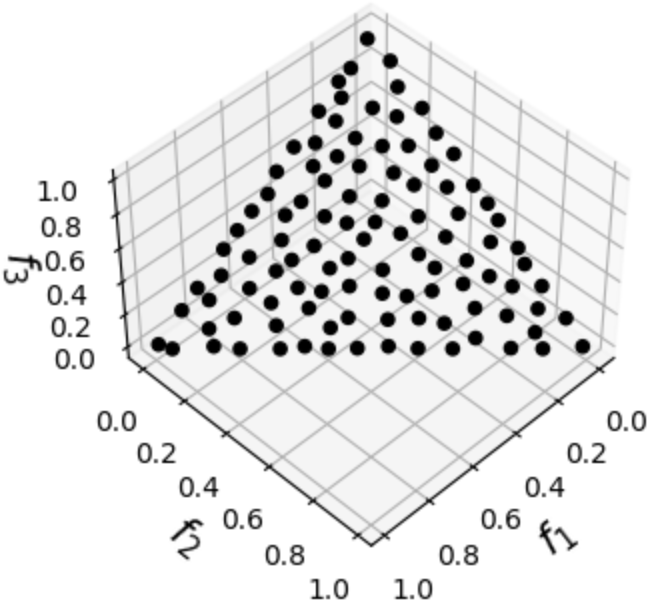} 
        \includegraphics[scale=0.155]{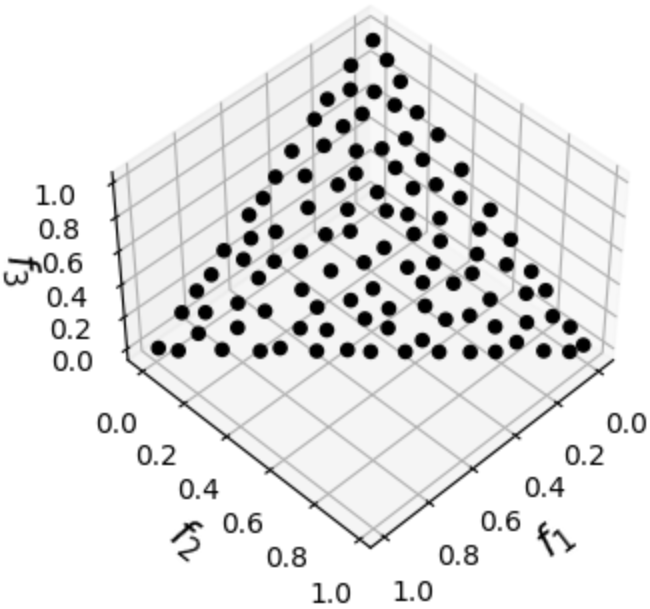} \\
       \makebox[2.765cm]{(a) \emph{immediate}} \makebox[2.765cm]{(b) \emph{batch}} \makebox[2.765cm]{(c) \emph{unbounded}} \\
       \makebox[2.765cm]{(IGD=0.0379)} \makebox[2.765cm]{(IGD=0.0389)} \makebox[2.765cm]{(IGD=0.0391)}
    \end{tabular}
    \end{center}
    \vspace{0pt}
    \begin{center}
    \begin{tabular}{@{}c@{\hspace{0pt}}c@{\hspace{0pt}}c@{}}
        \includegraphics[scale=0.155]{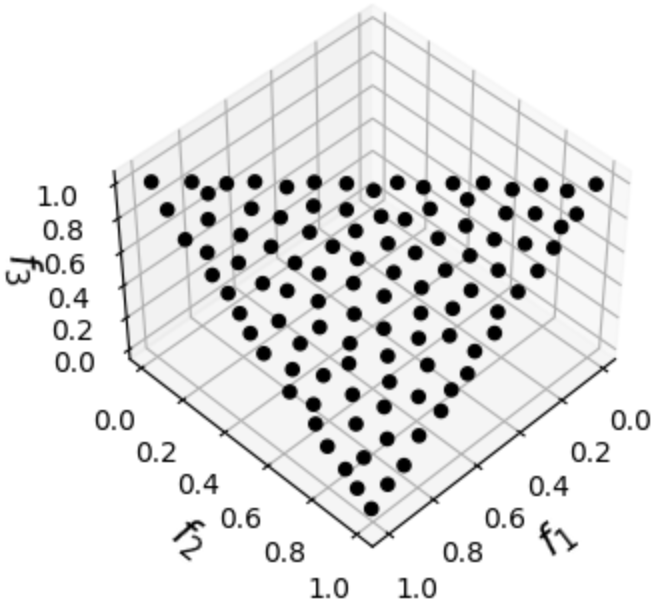} 
        \includegraphics[scale=0.155]{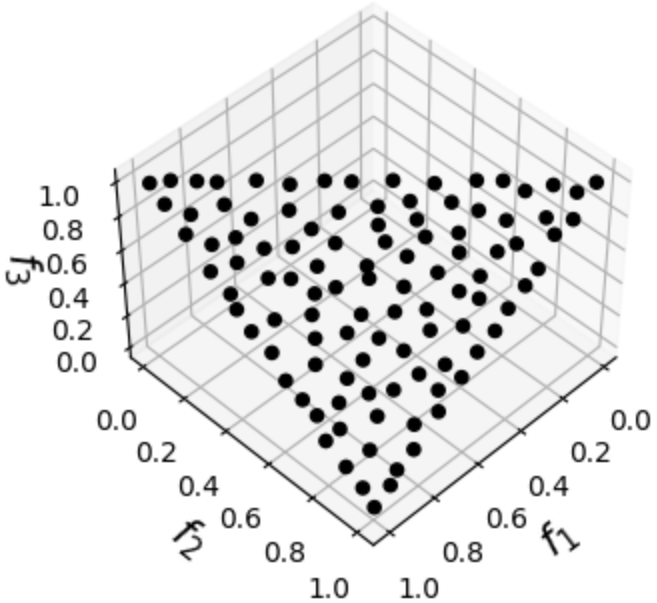} 
        \includegraphics[scale=0.155]{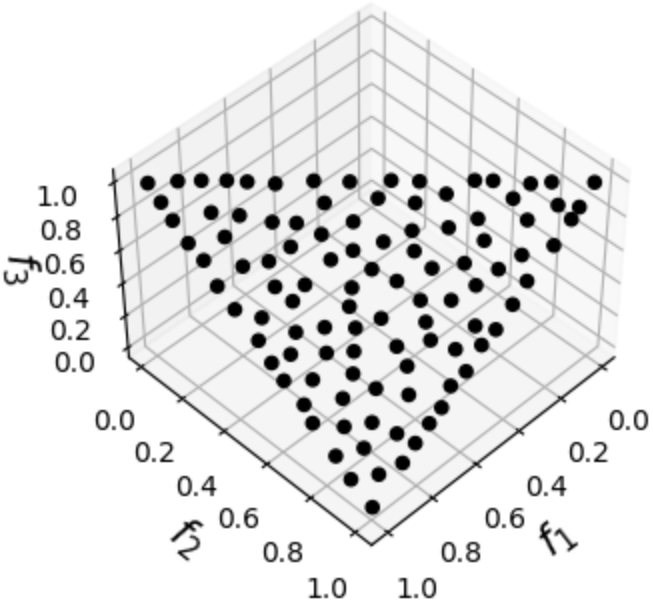} \\
       \makebox[2.765cm]{(a) \emph{immediate}} \makebox[2.765cm]{(b) \emph{batch}} \makebox[2.765cm]{(c) \emph{unbounded}} \\
       \makebox[2.765cm]{(IGD=0.0380)} \makebox[2.765cm]{(IGD=0.0390)} \makebox[2.765cm]{(IGD=0.0396)}
    \end{tabular}
    \end{center}
    \vspace{-10pt}
    \caption{Solutions (along with its IGD value) obtained by SMS-EMOA under the three truncation approaches on the simplex-shaped (top panel) and inverted simplex-shaped (bottom panel) sequences in the run with the median IGD value.}
    \label{fig4}
\end{figure}

SMS-EMOA uses the hypervolume indicator to truncate the archive iteratively. In each iteration, it calculates the hypervolume contribution of each solution to the set of candidate solutions and removes the solution with the least hypervolume contribution. The results obtained by the truncation method of SMS-EMOA are given in Table~\ref{tab:igd_results} and Figure \ref{fig4}. 
As we can see, the three truncation approaches can well maintain the solutions' uniformity, with their solutions being distributed fairly evenly on the Pareto front.

\begin{figure}[tbp]
    \begin{center}
    \begin{tabular}{@{}c@{\hspace{0pt}}c@{\hspace{0pt}}c@{}}
        \includegraphics[scale=0.155]{pictures/SMS-3.png} 
        \includegraphics[scale=0.155]{pictures/SMS-7.png} 
        \includegraphics[scale=0.155]{pictures/SMS-11.png} \\
       \makebox[2.765cm]{(a) \emph{immediate}} \makebox[2.765cm]{(b) \emph{batch}} \makebox[2.765cm]{(c) \emph{unbounded}} \\
       \makebox[2.765cm]{(IGD=0.0379)} \makebox[2.765cm]{(IGD=0.0389)} \makebox[2.765cm]{(IGD=0.0391)}
    \end{tabular}
    \end{center}
    \vspace{0pt}
    \begin{center}
    \begin{tabular}{@{}c@{\hspace{0pt}}c@{\hspace{0pt}}c@{}}
        \includegraphics[scale=0.155]{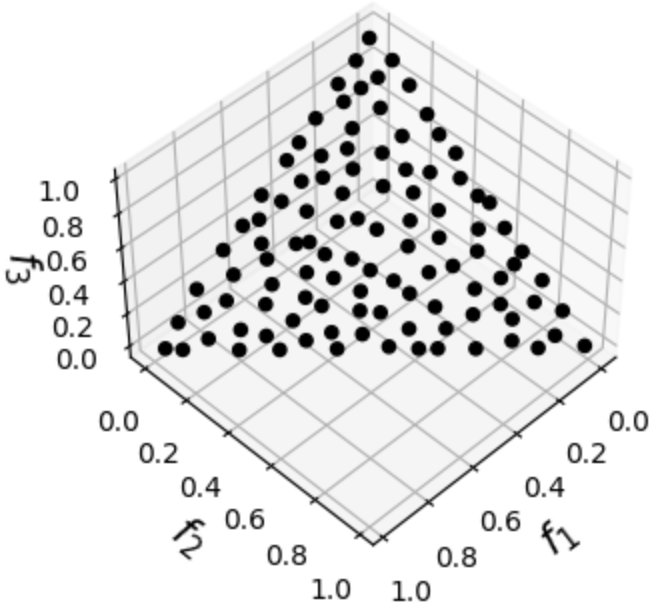} 
        \includegraphics[scale=0.155]{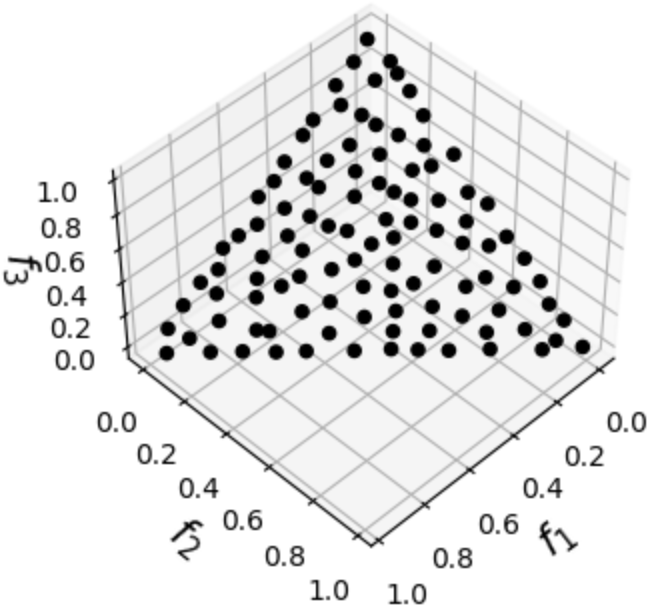} 
        \includegraphics[scale=0.155]{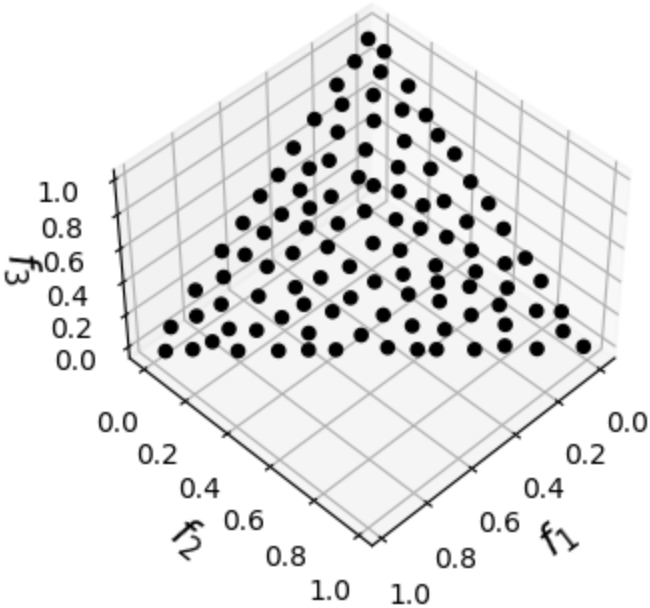} \\
       \makebox[2.765cm]{(a) \emph{immediate}} \makebox[2.765cm]{(b) \emph{batch}} \makebox[2.765cm]{(c) \emph{unbounded}} \\
       \makebox[2.765cm]{(IGD=0.0387)} \makebox[2.765cm]{(IGD=0.0392)} \makebox[2.765cm]{(IGD=0.0384)}
    \end{tabular}
    \end{center}
    \vspace{-10pt}
    \caption{Solutions (along with their corresponding IGD values) obtained by the original SMS-EMOA (i.e., removing the worst solutions; top panel) and the modified SMS-EMOA with the inclusion truncation method (i.e., including the best solutions; bottom panel) under the three truncation approaches on the simplex sequence.}
    \label{fig5}
\end{figure}




However, interestingly, the solutions obtained by the \emph{immediate} approach are the best with respect to their uniformity, followed by the \emph{batch} approach, and the worst is those obtained by the \emph{unbounded} approach, which can also be confirmed from their IGD values. 
This sounds a bit counter-intuitive since the \emph{unbounded} approach has all the solutions available, while the \emph{immediate} one can only see one-step ahead (i.e., only one solution available). One possible explanation is that despite the \emph{unbounded} approach having all the solutions, the truncation method in SMS-EMOA still uses the one-by-one greedy truncation\footnote{It is in general impossible to find the subset with the specified size that achieves the best quality with respect to a sub-modular function like hypervolume from a large set since it is an NP-hard problem~\cite{karp_reducibility_1972}.}, removing the solution contributing least to the set's hypervolume. Such a greedy truncation may introduce error (when the number of solutions to be truncated is larger than 1). In SMS-EMOA, for the \emph{unbounded} approach there are (5000 - 105) iterations, i.e., 4895 solutions needed to be removed. Consequently, the error will be accumulated and become pronounced after many iterations. 
This phenomenon is actually known as \textit{nesting effect} in the area of subset selection~\cite{pudil1994floating}. 
In contrast, the accumulated error should be smaller in the \emph{batch} approach, and even it is none in the \emph{immediate} approach. 

To further verify this, we now use an ``opposite'' way to truncate the archive in the \emph{unbounded} approach. That is, we still employ the one-by-one greedy method, but select the solution (into the archive) which has the largest hypervolume contribution to the current archive (rather than removing the solution which has the least hypervolume contribution to the set of candidate solutions, i.e., inclusion versus removal). Figure~\ref{fig5} gives the solutions obtained by this new hypervolume-based method (called \textit{inclusion}) for the \emph{unbounded} approach as well as the other two approaches. As can be seen, for the \emph{unbounded} approach the solutions obtained by the inclusion method perform slightly better than those by the original SMS-EMOA, while for the \emph{immediate} approach, the removal method performs better than the inclusion. This is due to different accumulated errors. For the \emph{unbounded} approach, there are 4895 solutions to be removed, whereas only 105 solutions needed to be included into the archive. For the \emph{immediate} approach, there is only one solution to be removed, whereas 105 solutions needed to be included. 

In addition, consider the results of the \emph{batch} approach shown in the middle of Figure~\ref{fig5}, where there are the same number of solutions (both 105) to be removed and included for the removal and inclusion methods, respectively. The removal is slightly better than the inclusion. This indicates that removing the most crowded solutions tends to be a better way than selecting the least crowded solutions when the number of solutions to be truncated is commensurate with the total solution number. This is probably a reason why most of existing MOEAs consider the removal other than the inclusion.

\subsection{IBEA}

\begin{figure}[tbp]
    \begin{center}
    \begin{tabular}{@{}c@{\hspace{0pt}}c@{\hspace{0pt}}c@{}}
        \includegraphics[scale=0.155]{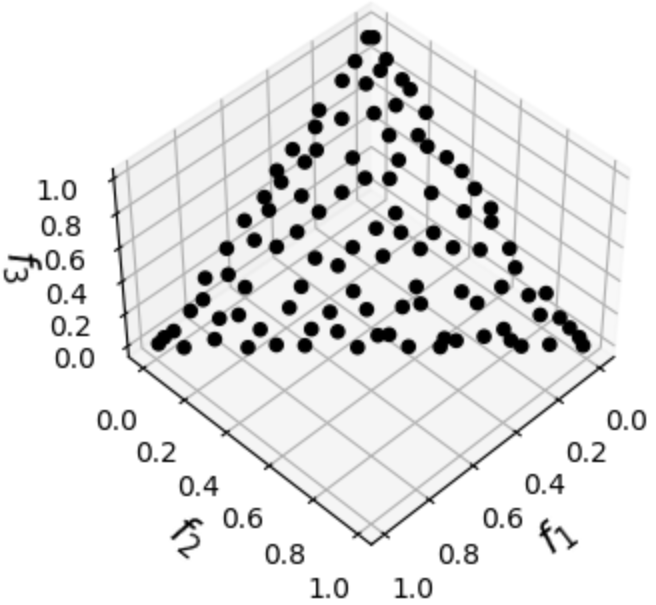} 
        \includegraphics[scale=0.155]{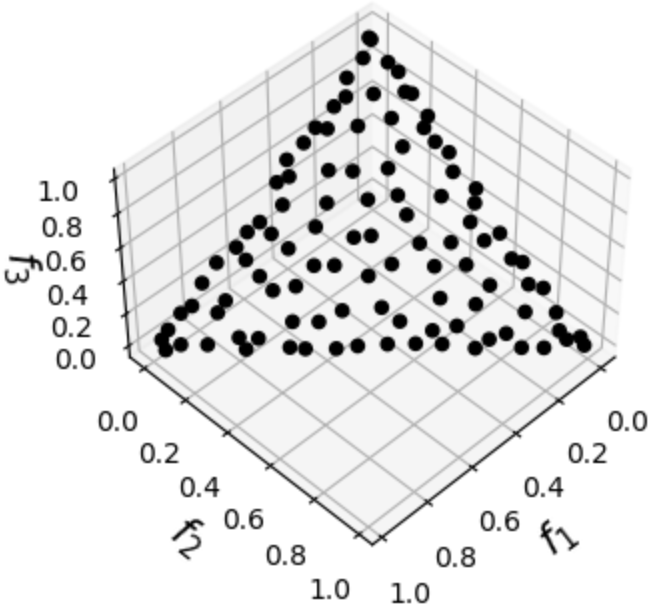} 
        \includegraphics[scale=0.155]{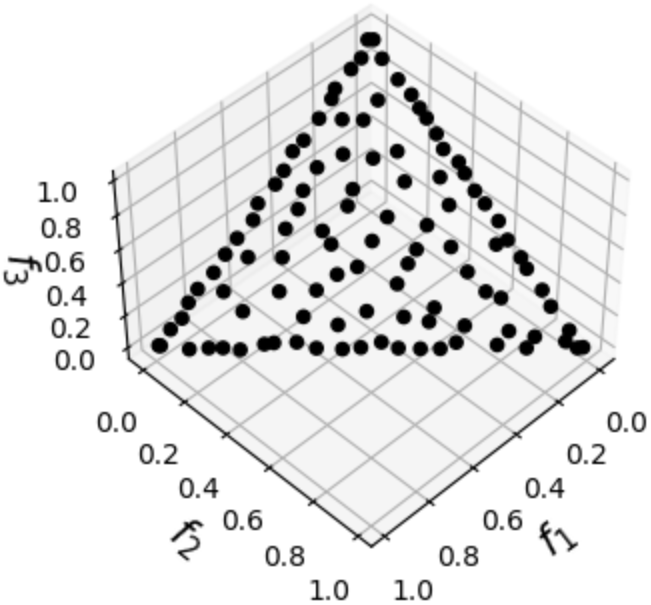} \\
       \makebox[2.765cm]{(a) \emph{immediate}} \makebox[2.765cm]{(b) \emph{batch}} \makebox[2.765cm]{(c) \emph{unbounded}} \\
       \makebox[2.765cm]{(IGD=0.0416)} \makebox[2.765cm]{(IGD=0.0414)} \makebox[2.765cm]{(IGD=0.0432)}
    \end{tabular}
    \end{center}
    \vspace{0pt}
    \begin{center}
    \begin{tabular}{@{}c@{\hspace{0pt}}c@{\hspace{0pt}}c@{}}
        \includegraphics[scale=0.155]{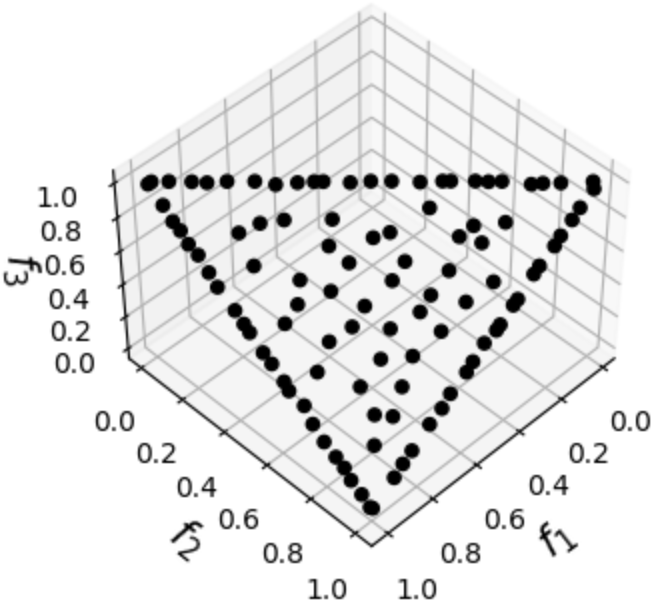} 
        \includegraphics[scale=0.155]{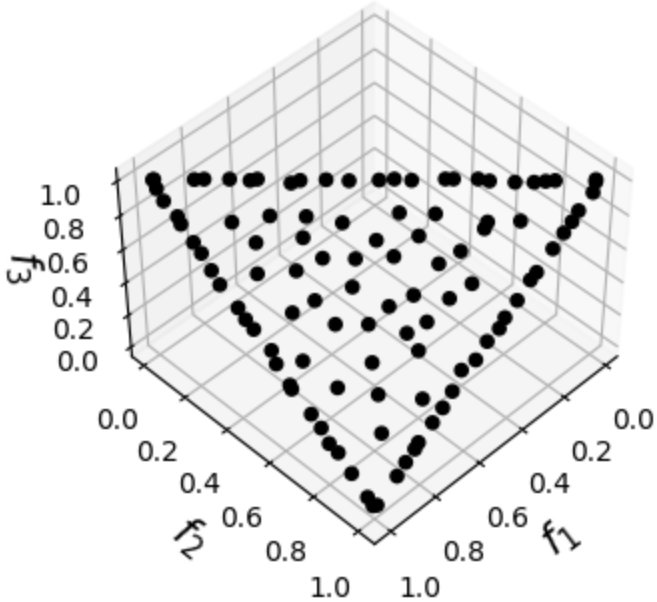} 
        \includegraphics[scale=0.155]{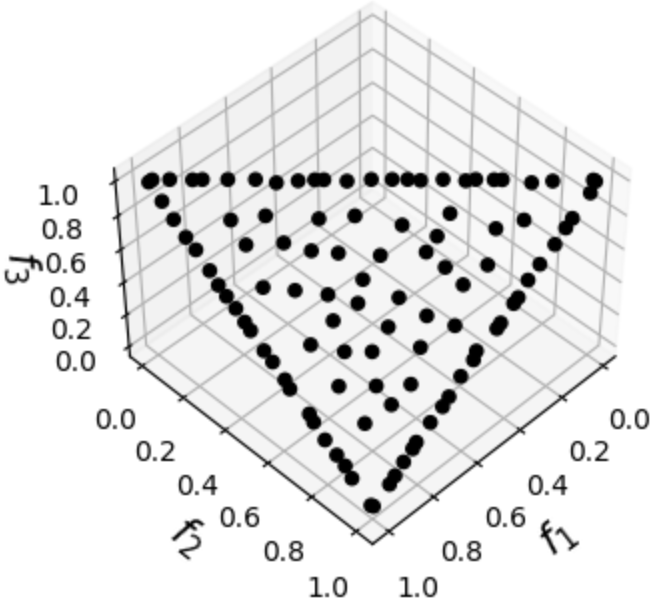} \\
       \makebox[2.765cm]{(a) \emph{immediate}} \makebox[2.765cm]{(b) \emph{batch}} \makebox[2.765cm]{(c) \emph{unbounded}} \\
       \makebox[2.765cm]{(IGD=0.0467)} \makebox[2.765cm]{(IGD=0.0463)} \makebox[2.765cm]{(IGD=0.0463)}
    \end{tabular}
    \end{center}
    \vspace{-10pt}
    \caption{Solutions (along with its IGD value) obtained by IBEA under the three truncation approaches on the simplex-shaped (top panel) and inverted simplex-shaped (bottom panel) sequences in the run with the median IGD value.}
    \label{fig6}
\end{figure}

Based on the same idea of SMS-EMOA, IBEA uses an indicator to measure the quality of a solution in the solution set. 
A major difference of IBEA to SMS-EMOA is that for a solution's indicator value, IBEA defines a transformation (exponential) function to do the pairwise comparison between the solution and any other solutions in the set, and add up the function values as the fitness of a solution. This may lead to the algorithm to preferring boundary solutions, as shown in \cite{Li2016,li2017modified}. 

Similar results are obtained under the three truncation approaches for both simplex and inverted simplex sequences (Figure~\ref{fig6}). 
It can be seen that the solutions are not distributed as uniformly as in SMS-EMOA, more crowded along the boundary lines of the Pareto front, particularly for the inverted simplex sequences. 


It is worth mentioning that the nesting effect observed in SMS-EMOA also applies to IBEA -- the one-by-one greedy truncation may accumulate larger error with more iterations. However, this seems to be largely overshadowed by the algorithm's behaviour of boundary solution preference. This may explain why unlike SMS-EMOA's clear performance degradation pattern over the increase of the archive size to be truncated from, IBEA shows less pronounced archive-size-dependent behaviour.

\subsection{MOEA/D}

\begin{figure}[tbp]
    \begin{center}
    \begin{tabular}{@{}c@{\hspace{0pt}}c@{\hspace{0pt}}c@{}}
        \includegraphics[scale=0.155]{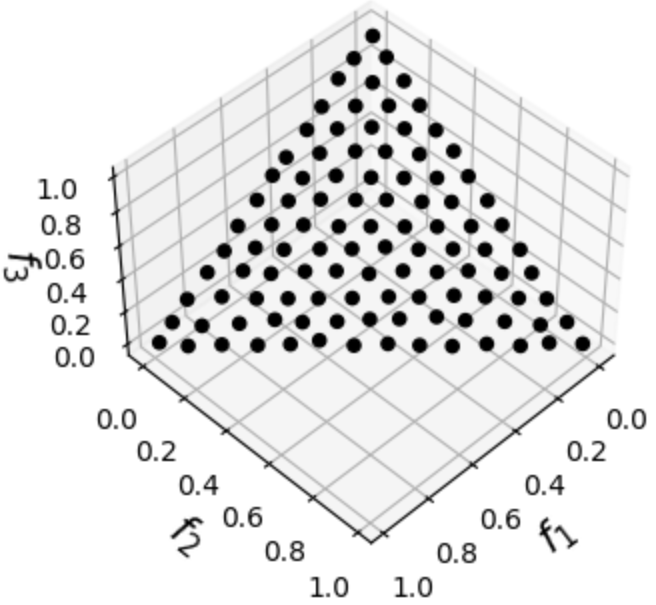} 
        \includegraphics[scale=0.155]{pictures/MOEAD-pbi-1.png} 
        \includegraphics[scale=0.155]{pictures/MOEAD-pbi-1.png} \\
       \makebox[2.765cm]{(a) \emph{immediate}} \makebox[2.765cm]{(b) \emph{batch}} \makebox[2.765cm]{(c) \emph{unbounded}} \\
       \makebox[2.765cm]{(IGD=0.0372)} \makebox[2.765cm]{(IGD=0.0372)} \makebox[2.765cm]{(IGD=0.0372)}
    \end{tabular}
    \end{center}
    \vspace{0pt}
    \begin{center}
    \begin{tabular}{@{}c@{\hspace{0pt}}c@{\hspace{0pt}}c@{}}
        \includegraphics[scale=0.155]{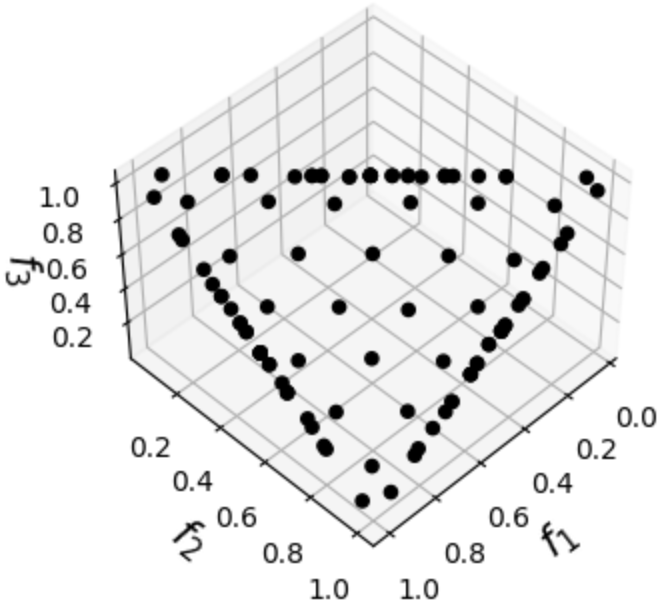} 
        \includegraphics[scale=0.155]{pictures/MOEAD-pbi-4.png} 
        \includegraphics[scale=0.155]{pictures/MOEAD-pbi-4.png} \\
       \makebox[2.765cm]{(a) \emph{immediate}} \makebox[2.765cm]{(b) \emph{batch}} \makebox[2.765cm]{(c) \emph{unbounded}} \\
       \makebox[2.765cm]{(IGD=0.0621)} \makebox[2.765cm]{(IGD=0.0621)} \makebox[2.765cm]{(IGD=0.0621)}
    \end{tabular}
    \end{center}
    \vspace{-10pt}
    \caption{Solutions (along with its IGD value) obtained by MOEAD under the three truncation approaches on the simplex-shaped (top panel) and inverted simplex-shaped (bottom panel) sequences in the run with the median IGD value.}
    \label{fig8}
\end{figure}

MOEA/D maintains the archive by decomposing a given multi-objective problem into a number of single-objective sub-problems using a set of uniformly distributed weights. Based on the scalarising function used, it selects the best solution to each sub-problem. 
The results of MOEA/D under the three truncation frequencies are given in Table \ref{tab:igd_results} and Figure \ref{fig8}. 
MOEA/D turns out to be the algorithm invariable to the truncation frequencies. 
As can be seen in the table, it has the same IGD means and zero standard deviation, meaning that the algorithm acquires the same solutions over all the 31 sequences under the three truncation approaches. 
Regarding the solution quality on the sequences with different shapes, it is well known that MOEA/D performs perfectly on a simplex Pareto front (as shown in the top panel of Figure \ref{fig8}), whereas it performs poorly on a Pareto front with an irregular shape like the inverted simplex (the bottom panel of Figure \ref{fig8}) \cite{Ishibuchi2016}.

\subsection{NSGA-III}

\begin{figure}[tbp]
    \begin{center}
    \begin{tabular}{@{}c@{\hspace{0pt}}c@{\hspace{0pt}}c@{}}
        \includegraphics[scale=0.155]{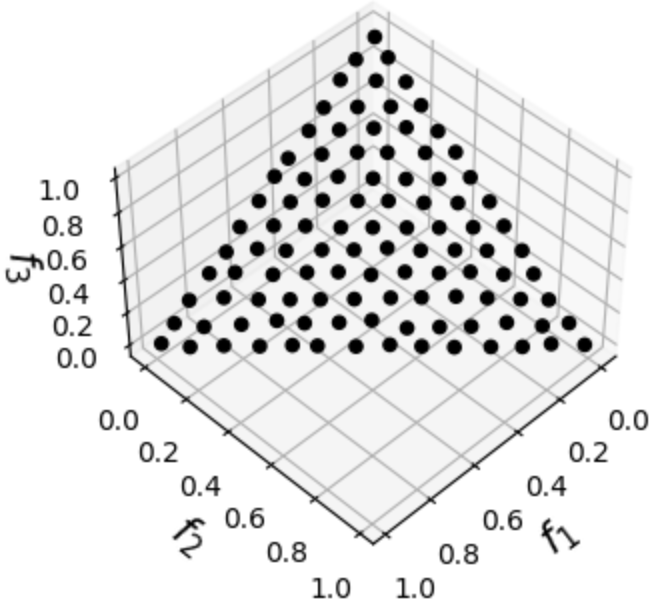} 
        \includegraphics[scale=0.155]{pictures/NSGA3-3.png} 
        \includegraphics[scale=0.155]{pictures/NSGA3-3.png} \\
       \makebox[2.765cm]{(a) \emph{immediate}} \makebox[2.765cm]{(b) \emph{batch}} \makebox[2.765cm]{(c) \emph{unbounded}} \\
       \makebox[2.765cm]{(IGD=0.0373)} \makebox[2.765cm]{(IGD=0.0373)} \makebox[2.765cm]{(IGD=0.0373)}
    \end{tabular}
    \end{center}
    \vspace{0pt}
    \begin{center}
    \begin{tabular}{@{}c@{\hspace{0pt}}c@{\hspace{0pt}}c@{}}
        \includegraphics[scale=0.155]{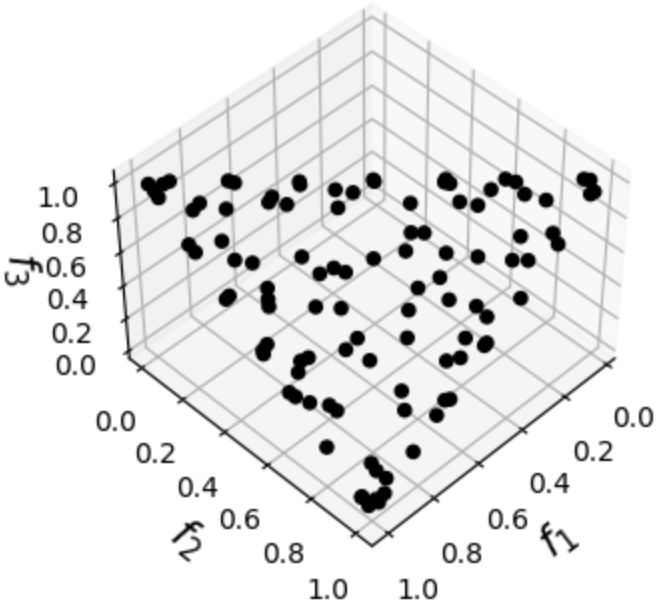} 
        \includegraphics[scale=0.155]{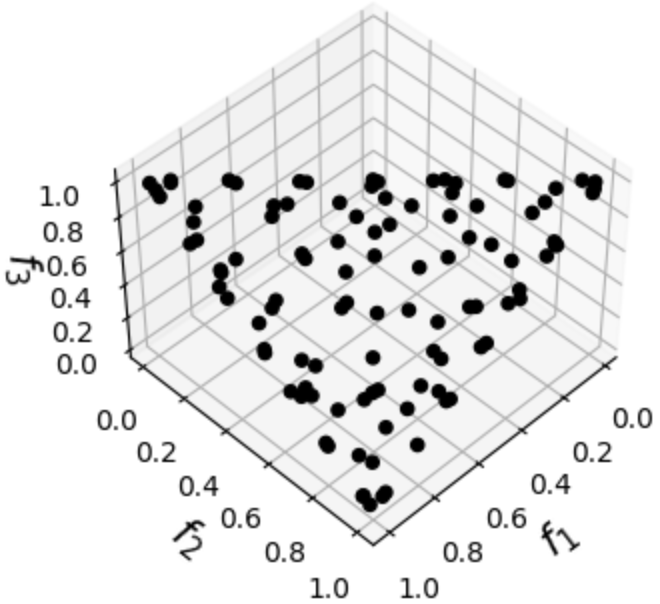} 
        \includegraphics[scale=0.155]{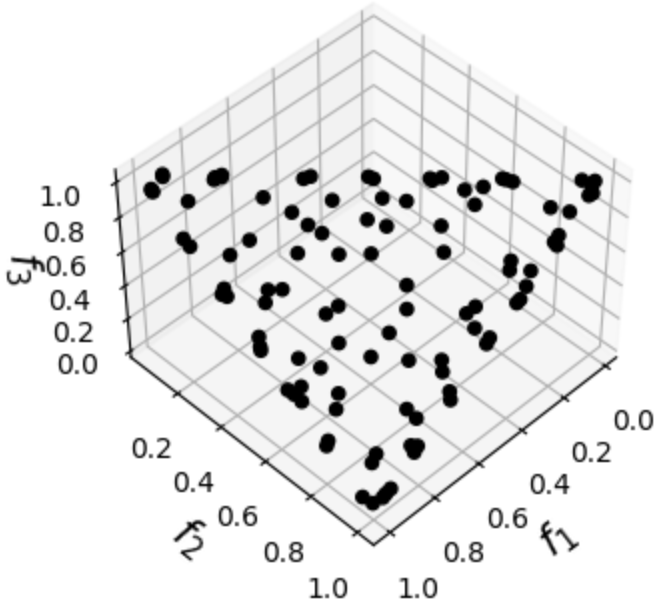} \\
       \makebox[2.765cm]{(a) \emph{immediate}} \makebox[2.765cm]{(b) \emph{batch}} \makebox[2.765cm]{(c) \emph{unbounded}} \\
       \makebox[2.765cm]{(IGD=0.0487)} \makebox[2.765cm]{(IGD=0.0494)} \makebox[2.765cm]{(IGD=0.0504)}
    \end{tabular}
    \end{center}
    \vspace{-10pt}
    \caption{Solutions (along with its IGD value) obtained by NSGA-III under the three truncation approaches on the simplex-shaped (top panel) and inverted simplex-shaped (bottom panel) sequences in the run with the median IGD value.}
    \label{fig9}
\end{figure}




Similar to MOEA/D, NSGA-III also relies on a set of reference points (weight vectors) to perform the truncation. However, one difference is that NSGA-III uses the reference points to balance the convergence and diversity by associating solutions with the closest reference point and selecting solutions based on the non-dominated sorting (convergence) and niche selection (diversity) criteria. When there is a mismatch where there are more than one solution associated with a reference point, after choosing the closest one, NSGA-III randomly chooses others to fill the slots left (if any). 

The results of NSGA-III under the three truncation frequencies are given in Table \ref{tab:igd_results} and Figure \ref{fig9}. As can be seen, on the simplex-shape sequence, all the three truncation approaches perform very similarly, with an identical mean IGD value and the same solutions obtained by the run with median IGD. The different result comes from the case with the inverted simplex-shape sequence -- the \emph{immediate} approach has the best IGD, followed by the \emph{batch} approach, and the worst is the \emph{unbounded} approach; the difference is statistically significant.   
This is unexpected since NSGA-III randomly chooses solutions associated with the weight vectors after the closest match, hence differences of the solutions under the three truncation frequencies are likely to be statistically indifferent. 
A possible explanation for this may be attributed to the normalisation mechanism used in NSGA-III. The algorithm employs the ideal point and nadir point of the current solutions to determine the reference points. In the \emph{immediate} approach, the reference points may change before the global ideal and nadir points found, which may help keep diverse solutions (as different reference points correspond to different solutions) to some extent (i.e., better than pure random selection).
That said, further examinations are needed to confirm our conjecture and to understand the archiving behaviour of NSGA-III under different truncation frequencies.

\section{Limitations}

In this study, we consider the sequences consisting of solutions randomly sampled from the Pareto front, i.e., they are nondominated to each other. 
Another way is to consider the sequences obtained by an MOEA, e.g., initial solutions, offspring solutions in the 1st generation, offspring solutions in the 2nd generation, etc. 
We expect that the second way may obtain different results as the Pareto dominance relation will play a role here for the \emph{immediate} and \emph{batch} approaches -- the archive may deteriorate, i.e., it may accept solutions which are dominated by the solutions removed previously, as the truncation methods in the considered MOEAs are all subject to deterioration \cite{li2024multi}. We will leave this for future study.

The archive used in this study is not involved in the solution generation process, i.e., it is only used as an external archive. 
If an ``internal'' archive is used in MOEAs, considering an unbounded archive can be more beneficial since it stores all the best solutions found as parents to generate new solutions. In contrast, the \emph{immediate} and \emph{batch} approaches need to remove some, despite the fact that they should typically be significantly faster than the \emph{unbounded} one.  
On the other hand, when the archive is involved in the solution generation process, it may not always beneficial to preserve the best solutions -- solutions that have high potential to generate promising offspring may be those with relatively low quality but can serve as a ``bridge'' to connect multiple local optimal regions. Preserving such solutions can benefit the search, which has been shown empirically \cite{tanabe2019non,Liang2023} and analytically \cite{bian2025stochastic,ren2025stochastic}.  

It is worth mentioning that in this study we used the 3-objective cases with linear Pareto fronts. Different objective numbers and Pareto front shapes can certainly have different effects to archiving methods. For example, decomposition-based methods may struggle on irregular Pareto fronts \cite{Ishibuchi2016} and Pareto-based methods may fail on many-objective problems \cite{Wagner2007,Li2013b}. 
However, since the aim of this study is to investigate the effect of the truncation frequency of archiving methods to their performance, we want to choose simple cases that can typically be easy to be dealt with, hence minimising the effect of other factors rather than the truncation frequency.     




\section{Conclusion}

This paper investigated the effect of the timing of truncating the archive: 1) once a new solution arrives (\emph{immediate} approach), 2) once a batch of new solutions arrive (\emph{batch} approach), and 3) after all solutions arrive (\emph{unbounded} approach). We considered the truncation methods from five representative MOEAs. Our findings are summarised as follows.

\begin{itemize}
    \item The one-off truncation method which does not update the solution information (as in NSGA-II) is not suitable for truncating a large archive (e.g., in the \emph{unbounded} approach).

    \item The one-by-one greedy removal method which removes the worst solutions iteratively (used in SMS-EMOA and IBEA) is not suitable for the \emph{unbounded} approach.

    \item Truncating the archive by removal tends to perform slightly better than by inclusion (i.e., adding the best solutions iteratively into the archive) when the number of solutions to be truncated down to is commensurate with the total solution number (e.g., in the \emph{batch} approach). 
    
    \item For many MOEAs (NSGA-II, SMS-EMOA and IBEA), the \emph{immediate} approach is in general the best, followed by the \emph{batch}, and the \emph{unbounded} is the worst; while for some other MOEAs (MOEA/D and NSGA-III), the effect of the timing of truncating the archive is minor, if any. 
\end{itemize}

The reason for the last, seemingly counter-intuitive findings, is the combination of a large archive and the use of the one-by-one solution removal strategy. It takes $N-\mu$ operations for the removal strategy to reduce the archive down to the required size $\mu$ (where $N$ is the number of solutions in the archive waiting to be truncated), leading to a substantial accumulated error.

In addition, the results shown in this paper also suggest a need of developing effective truncation methods when considering unbounded archiving. Simply using what have been used in the population maintenance procedure of MOEAs may not work since the size to be reduced down to is very small compared to the total number of nondominated solutions in the archive. Encouragingly, 
researchers in the field have begun making efforts to address this issue (e.g., \cite{bringmann2014generic,singh2018distance,nan2022improved,chen2021fast,gu2023subset,wang2022enhancing}). 


\bibliography{sample-base}
\bibliographystyle{unsrt}

\appendix

\end{document}